\documentclass[letterpaper, 10 pt, conference]{ieeeconf}  

\IEEEoverridecommandlockouts                             

\usepackage[colorlinks]{hyperref}
\usepackage[dvipdfm]{graphicx}
\usepackage{color}
\usepackage[table,dvipsnames]{xcolor}
\usepackage{amsmath,amssymb}
\usepackage{caption}
\usepackage{epstopdf}
\usepackage{rotating}
\usepackage{cite}
\usepackage{wrapfig}
\usepackage{float}
\usepackage{titlesec}
\usepackage{algorithm}
\usepackage{algpseudocode}
\usepackage{array}
\usepackage{booktabs}
\usepackage{makecell, cellspace}
\definecolor{blond}{rgb}{0.98, 0.98, 0.82}

\allowdisplaybreaks

\titlespacing{\section}{0pt}{2mm}{1mm}
\titlespacing{\subsection}{0pt}{1mm}{0.25mm}
\titlespacing{\subsubsection}{0pt}{0mm}{0mm}

\renewcommand{\baselinestretch}{0.93}

\title{\LARGE \bf
CoBL-Diffusion: Diffusion-Based Conditional Robot Planning in Dynamic Environments Using Control Barrier and Lyapunov Functions
}

\author{Kazuki Mizuta$^1$ $\quad$ Karen Leung$^{1,2}$
\thanks{Kazuki Mizuta is partially supported by the Nakajima Foundation. This work was supported by the UW + Amazon Science Hub Research Award.}
\thanks{$^1$University of Washington, Department of Aeronautics and Astronautics, $^2$ NVIDIA,
        Contact: {\tt\small \{mizuta, kymleung\}@uw.edu }}
\thanks{}
}

\newcommand{\algo}{CoBL-Diffusion}

\usepackage{cite}
\usepackage{url}

\usepackage{amsmath} 
\usepackage{amssymb}  
\usepackage{mathtools} 
\usepackage{enumerate}

\usepackage{silence}
\usepackage{subcaption}
\usepackage{cleveref}
\crefname{equation}{}{} 
\crefname{section}{}{} 
\crefname{figure}{}{} 
\crefname{table}{}{} 
\crefname{lemma}{}{} 
\crefname{assumption}{}{} 
\crefname{remark}{}{} 
\crefname{corollary}{}{} 
\crefname{theorem}{}{} 
\crefname{problem}{}{} 
\crefrangeformat{equation}{(#3#1#4) to~(#5#2#6)} 
\crefformat{footnote}{#2\footnotemark[#1]#3} 

\let\originalleft\left
\let\originalright\right
\renewcommand{\left}{\mathopen{}\mathclose\bgroup\originalleft}
\renewcommand{\right}{\aftergroup\egroup\originalright}

\newtheorem{theorem}{Theorem}

\newtheorem{definition}{Definition}

\def\QED{\mbox{\rule[0pt]{1.5ex}{1.5ex}}}

\renewcommand{\exp}[1]{\ensuremath{\textrm{exp}\para{#1}}}

\renewcommand{\log}{\ensuremath{\textrm{log}}}

\newcommand{\para}[1]{\left( #1 \right)}

\newcommand{\ul}[1]{\underline{#1}} 

\begin{document}

\maketitle
\thispagestyle{empty}
\pagestyle{empty}

\begin{abstract}
Equipping autonomous robots with the ability to navigate safely and efficiently around humans is a crucial step toward achieving trusted robot autonomy.
However, generating robot plans while ensuring safety in dynamic multi-agent environments remains a key challenge.
Building upon recent work on leveraging deep generative models for robot planning in static environments, this paper proposes \algo, a novel diffusion-based safe robot planner for dynamic environments. \algo~uses \ul{Co}ntrol \ul{B}arrier and \ul{L}yapunov functions to guide the denoising process of a diffusion model, iteratively refining the robot control sequence to satisfy the safety and stability constraints. 
We demonstrate the effectiveness of \algo~using two settings: a synthetic single-agent environment and a real-world pedestrian dataset.
Our results show that \algo~generates smooth trajectories that enable the robot to reach goal locations while maintaining a low collision rate with dynamic obstacles.
\end{abstract}

\section{Introduction}

Achieving safe and efficient navigation for autonomous robots around humans is essential for building trust in autonomous systems and enabling their widespread adoption.
As ensuring safety is paramount, especially when interacting with humans, it is often desirable to leverage control-theoretic tools such as reachability analysis \cite{AlthoffDolan2014,AlthoffFrehseEtAl2021} and control invariant set theory \cite{AmesCooganEtAl2019,MitchellBayenEtAl2005} to define and enforce safety constraints within a model-based trajectory optimizer. 
While such approaches produce well-understood and well-behaved robot motions \cite{LeungSchmerlingEtAl2020,KousikHolmesEtAl2019,ChenTomlin2018}, they quickly become overly conservative and/or intractable for uncertain and complicated settings.

In contrast, emerging data-driven, i.e., deep learning, robot planners provide a scalable and tractable solution as they can leverage data from simulation or real-world interactions \cite{RhinehartMcAllisterEtAl2020,IvanovicLeungEtAl2020} and use high-dimensional observations as inputs.
However, learning-based approaches inherently lack safety guarantees and can exhibit unpredictable failures, leading to unanticipated behaviors that pose safety risks for robots and their surroundings.
Recent work has investigated applying control-theoretic safety filters on learning-based planners to combine the safety and generalizability benefits from each method \cite{XiaoWangEtAl2023b,YangChenEtAl2022,XiaoWangEtAl2022,PhanMinhHowingtonEtAl2023}.
In this work, we study further the advantages of integrating control-theoretic techniques for safety assurance into diffusion-based motion planning frameworks.

Diffusion models \cite{SohlDicksteinWeissEtAl2015, HoJainEtAl2020} have emerged as a powerful class of deep generative models, achieving remarkable success in generating high-quality images and synthesizing speech \cite{RombachBlattmannEtAl2022, HuangZhaoEtAl2022}, and more recently, in robot trajectory generation \cite{JannerDuEtAl2022, ZhongRempeEtAl2023, MishraXueEtAl2023, JiangCornmanEtAl2023}. 
However, current Diffusion-based robot trajectory generation methods still lack explicit safety assurances in dynamic environments, prohibiting their application in safety-critical scenarios involving human-robot interaction.
In this work, we investigate the use of diffusion models as the foundation for robot navigation planning in dynamic multi-agent environments and present a method that integrates control-theoretic safety and stability constraints into the diffusion model framework.
\begin{figure}[t]
  \centering
  \includegraphics[width=0.9\linewidth]{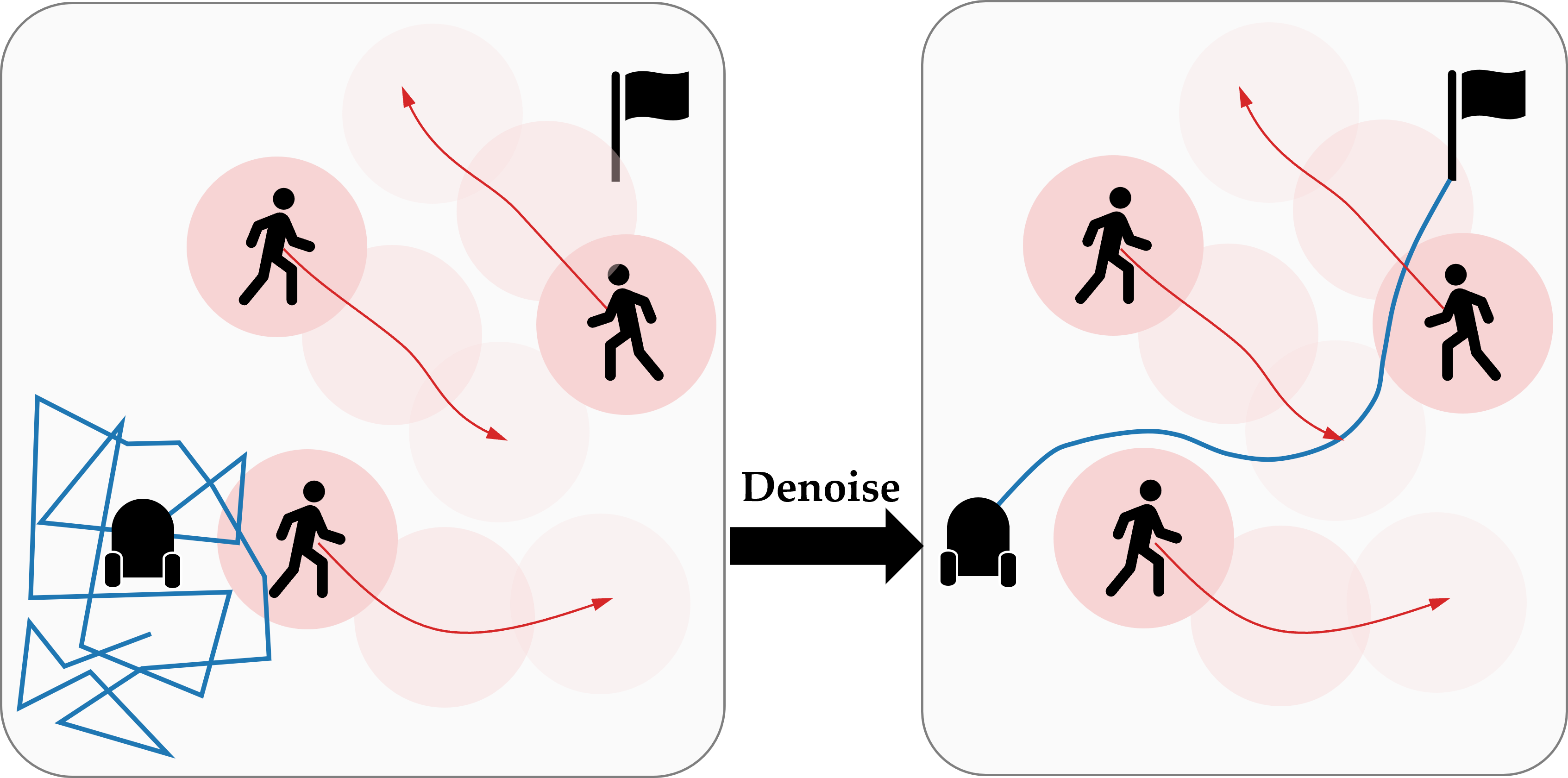}
  \caption{CoBL-Diffusion uses control barrier and Lyapunov functions to guide a diffusion process to generate a robot controller for goal-reaching while avoiding dynamic obstacles.}
  \label{fig:overview}
  \vspace{-3mm}
\end{figure}

Inspired by the Diffuser introduced in \cite{JannerDuEtAl2022} which incorporates user-defined reward functions into the diffusion process for flexible conditioning on the output behavior, we propose \textbf{\algo}, a novel diffusion model for safe robot planning leveraging \textbf{Co}ntrol \textbf{B}arrier Functions (CBFs) and Control \textbf{L}yapunov Functions (CLFs) for the enforcement of desired safety and stability (i.e., goal-reaching) properties.
CBFs and CLFs are popular and well-understood mathematical frameworks based on invariant set theory for proving and ensuring the safety and stability of dynamical systems \cite{AmesCooganEtAl2019}.
They have been successfully applied to mobile robots for safe collision-free navigation \cite{SrinivasanDabholkarEtAl2020, KhanIbukiEtAl2022, MizutaHirohataEtAl2022}. 
In this paper, we present the architecture behind \algo ~that generates dynamically feasible trajectories that satisfy the CBF and CLF constraints.
We demonstrate and evaluate the performance of \algo ~in navigation tasks using synthetically generated and real-world scenarios, showing that it produces robot motions that can safely and smoothly navigate through crowded human environments. 

\noindent \textbf{Statement of Contributions.} The contributions of the paper are summarized as follows:
\begin{enumerate}
    \item We propose \textbf{\algo}, a diffusion-based robot planner for dynamic environments that incorporates control-theoretic safety and stability constraints.
    The diffusion process is guided by the gradient of reward functions derived from the CBF and CLF.
    \item Our method ensures consistency between the control sequence and the resulting states by explicitly integrating generated controls through system dynamics to obtain the states.
    Maintaining dynamic consistency during the diffusion steps is crucial, as the CBF and CLF reward computations in \algo ~require the state trajectory to be consistent with the control inputs.
    \item We demonstrate the effectiveness of our approach in generating safe robot motion plans for navigating through crowded dynamic environments. 
    We provide qualitative and quantitative evaluations using synthetic and real-world pedestrian data.
\end{enumerate}

\noindent \textbf{Organization.} We first contextualize our work in Section~\ref{sec:related_work}, and then introduce key concepts on diffusion models, CBFs, and CLFs in Section~\ref{sec:preliminaries}.
In Section~\ref{sec:proposed_model}, we describe \algo, including the architecture and choice of reward functions.
We discuss experimental results of \algo ~evaluated on a synthetic and real-world dataset in Section~\ref{sec:experiment}, and conclude with ideas for future work in Section~\ref{sec:conclusion}.

\section{Related work} \label{sec:related_work}
In this section, we provide a brief overview of robot planning algorithms using CBFs and diffusion models.

\noindent\textbf{Control Barrier Functions (CBFs).} CBFs provide an elegant mathematical framework based on invariant set theory to construct safe control constraints that prevent a dynamical system from entering an unsafe region in state space \cite{AmesCooganEtAl2019}. 
A brief introduction on CBF is given in Section~\ref{subsec:CBF}.
CBFs have been successfully applied to various robotic applications, ranging from navigation \cite{SrinivasanDabholkarEtAl2020, LongQianEtAl2021}, robot manipulation \cite{CortezOetomoEtAl2019}, and legged locomotion \cite{GrandiaTaylorEtAl2021}.
For high-dimensional settings (e.g., multi-agent, history-dependence) and when operating with uncertain sensor observations, there has been a recent surge in combining CBFs with learning-based techniques, including learning CBFs from data \cite{YuHirayamaEtAl2023,QinZhangEtAl2021}, and integrating CBFs as a safety filter for learning-based controllers \cite{PereiraWangEtAl2021,XiaoWangEtAl2023b,WabersichTaylorEtAl2023}.
In general, there is a growing interest in using CBFs as a safety filter for learning-enabled systems that otherwise would struggle with confidently exhibiting safe behaviors.
In this work, we investigate the integration of CBFs into diffusion models, a recent generative modeling technique that is showing promise for robot planning applications \cite{JannerDuEtAl2022,XiaoWangEtAl2023,ZhongRempeEtAl2023}.
\noindent \textit{Note}: There are strong parallels between CBFs and CLFs; see Section~\ref{subsec:CLF}. Many CBF-based approaches discussed also apply to CLFs, which are used for stability \cite{DawsonGaoEtAl2022,ChangRoohiEtAl2019}. 

\noindent\textbf{Diffusion Models for Planning.} Diffusion models, introduced by \cite{SohlDicksteinWeissEtAl2015} and further developed by \cite{HoJainEtAl2020}, are a class of deep generative models that represent data generation as an iterative denoising process.
We provide a brief overview of diffusion models in  Section~\ref{subsec:ddpm}.
The Diffuser \cite{JannerDuEtAl2022} represents an innovative application of diffusion models to robot trajectory generation.
The Diffuser combines the planning and sampling processes into a unified framework via iterative denoising of trajectories, thus enabling the generation of plans that are adaptable to various reward structures and constraints.
There are follow-up works that aim to influence the diffusion process to produce trajectories that satisfy some desired properties.
Work similar in spirit to our proposed method, SafeDiffuser \cite{XiaoWangEtAl2023} extends the application of diffusion models to safety-critical tasks by incorporating CBFs into the denoising diffusion procedure to ensure the generated trajectories satisfy safety constraints.
This CBF guarantees forward invariance of the diffusion procedure rather than the conventional forward invariance in a time domain addressed in this work.
Another recent work \cite{ZhongRempeEtAl2023} proposes guided conditional diffusion in the context of simulating human driving behaviors in driving simulators. 
They use signal temporal logic specifications to encode desired behaviors such as goal-reaching and staying within speed limits.
\section{Preliminaries}\label{sec:preliminaries}

Our approach consists of a combination of control-theoretic techniques and deep generative models to generate robot behaviors with desired safety and stability properties.
In this section, we provide background on (i) planning with diffusion models, (ii) control barrier functions (CBFs), and (iii) control Lyapunov functions (CLFs).

\subsection{Diffusion Models}
\label{subsec:ddpm}
Deep generative modeling is the process of using deep neural networks to learn an underlying distribution explaining a dataset and sampling from the learned distribution to produce new examples similar to the dataset.
A denoising diffusion probabilistic model (DDPM) \cite{HoJainEtAl2020} is a type of deep generative model consisting of two main processes: \textit{forward diffusion} which turns clean data into white noise, and \textit{reverse diffusion} which turns white noise into clean data (i.e., denoising process). By learning neural network parameters describing the reverse diffusion process, we can generate clean data samples by sampling from, typically, a Gaussian distribution.
The forward process is defined by a sequence of conditional Gaussian distributions:
\begin{align} \label{eq:forward_diff}
    q(\boldsymbol{\tau}^i|\boldsymbol{\tau}^{i-1}) = \mathcal{N}(\boldsymbol{\tau}^i;\sqrt{1-\beta_i}\boldsymbol{\tau}^{i-1}, \beta_i \mathbf{I}),
\end{align}
where $\boldsymbol{\tau}^0$ is the original data, $\boldsymbol{\tau}^i$ is the data at diffusion step $i$, $\beta_i$ is the noise schedule, and $\mathbf{I}$ is the identity matrix.
Intuitively, the forward diffusion process incrementally adds noise to the data, transforming the original distribution into a simple Gaussian distribution after $N$ steps.
The reverse process is modeled by a series of conditional distributions:
\begin{align} \label{eq:reverse_diff}
    p_\theta(\boldsymbol{\tau}^{i-1}|\boldsymbol{\tau}^i) = \mathcal{N} (\boldsymbol{\tau}^{i-1};\boldsymbol{\mu}_\theta(\boldsymbol{\tau}^i, i), \mathbf{\Sigma}_\theta(\boldsymbol{\tau}^i,i)),
\end{align}
where $\boldsymbol{\mu}_\theta$ and $\mathbf{\Sigma}_\theta$ are parameterized by neural networks with parameters $\theta$.
In this paper, we set $\mathbf{\Sigma}_\theta(\boldsymbol{\tau}^i,i) = \mathbf{\Sigma}^i$ for the reverse diffusion process to follow the cosine schedule \cite{NicholDhariwal2021}.
The goal is to learn neural network parameters $\theta$ such that the reverse process reproduces the original data.
The typical training loss uses a variant of the variational lower bound on the log-likelihood,
\begin{align} \label{eq:diff_loss}
    L_\mathrm{simple} = \mathbb{E}_{\boldsymbol{\tau}^0, \boldsymbol{\epsilon}, i}[\|\boldsymbol{\epsilon}-\boldsymbol{\epsilon}_\theta(\sqrt{\Bar{\alpha}_i}\boldsymbol{\tau}^0+\sqrt{1-\Bar{\alpha}_i}\boldsymbol{\epsilon}, i)\|^2],
\end{align}
where $\boldsymbol{\epsilon}$ is Gaussian noise, $\boldsymbol{\epsilon}_\theta$ is the neural network predicting the noise, and $\Bar{\alpha}_i=\prod_{j=1}^i(1-\beta_j)$ is the cumulative product of the noise schedule.
To generate new samples, one starts from noise and iteratively applies \eqref{eq:reverse_diff} for a predetermined number of iterations. 

\subsection{Trajectory Generation Using Diffusion Models}
\label{subsec:diffuser}

Diffusion models have demonstrated their potential in various planning and trajectory generation problems.
Specifically, Diffuser \cite{JannerDuEtAl2022} reformulates the trajectory optimization by integrating the planning process into the modeling framework based on a diffusion model, thus providing the underlying foundation that this (and many other) work builds upon.
Unlike traditional algorithms that generate state sequences (i.e., trajectories) autoregressively, Diffuser generates all timesteps of a plan simultaneously, allowing for anti-causal decision-making and temporal coherence, thereby promoting the global consistency of the sampled plan.
In Diffuser, the trajectory consists of both states $\mathbf{s}$ and actions $\mathbf{a}$\footnote{In \cite{JannerDuEtAl2022}, they use $\mathbf{s}, \mathbf{a}$ for state and action (equivalently control) as that is typical notation for the reinforcement learning community. In this work, we use $\mathbf{x}, \mathbf{u}$ for state and control which is typical notation in the controls community.}, 
\begin{align}\label{eq:diffuser_traj}
    \boldsymbol{\tau} = 
    \begin{bmatrix}
        s_0\ s_1\ \ldots\ s_T\\
        a_0\ a_1\ \ldots\ a_T
    \end{bmatrix}.
\end{align}
The optimality of a state-action pair is denoted by $O_t$, a binary random variable where $p(O_t=1)=\exp{r(s_t, a_t)}$ and $r(s_t, a_t)$ is a user-specified reward function.
To influence the denoising process to generate more optimal outputs with respect to $r(s_t, a_t)$, the reward function serves as a classifier guidance \cite{DhariwalNichol2021} to modulate the mean in the denoising steps,
\begin{align}\label{eq:diffuser_reverse_diffusion}
    p_\theta(\boldsymbol{\tau}^{i-1}|\boldsymbol{\tau}^{i}, O_{1:T})\approx \mathcal{N}(\boldsymbol{\tau}^{i-1};\boldsymbol{\mu}+\mathbf{\Sigma} \mathbf{g}, \mathbf{\Sigma}),\\
    \mathbf{g} = \nabla_{\boldsymbol{\tau}} \log\ p(O_{1:T}|\boldsymbol{\tau})
    = \sum_{t=0}^T \nabla_{\mathbf{s}_t, \mathbf{a}_t} r(\mathbf{s}_t, \mathbf{a}_t),
\end{align}
where $\mathbf{g}$ represents the gradient of the log probability of optimality with respect to the trajectory.
Furthermore, the trajectory is conditioned on a given goal by fixing the terminal state, analogous to the inpainting problem from image generation \cite{SohlDicksteinWeissEtAl2015}.

\subsection{Control Barrier Function (CBF)} 
\label{subsec:CBF}
Given an unsafe set to avoid, CBFs impose a bound on the rate at which a dynamical system is allowed to approach the unsafe set, enforcing a rate of zero at the boundary of the unsafe set.
Consider a control affine robotic system,
\begin{align}\label{eq:dynamics}
    \Dot{\mathbf{x}}=f(\mathbf{x})+g(\mathbf{x})\mathbf{u}, 
\end{align}
where $\mathbf{x}\in \mathcal{D}\subseteq\mathbb{R}^n$ is the state, $\mathbf{u} \in \mathcal{U} \subseteq \mathbb{R}^m$ is the control input.
Both $f:\mathcal{D}\rightarrow\mathbb{R}^n$ and $g:\mathcal{D}\rightarrow\mathbb{R}^{n\times m}$ are local Lipschitz continuous functions.
Suppose that the safe set $\mathcal{C}\subseteq\mathcal{D}$ is defined as the superlevel set of the continuously differentiable function $h:\mathcal{D}\to\mathbb{R}$ as follows,
\begin{align}
    \mathcal{C} = \{\mathbf{x} \in  \mathcal{D} \mid h(\mathbf{x})\geq 0\}. \label{eq:safe_set}
\end{align}

\begin{definition}[Control Barrier Function \cite{AmesCooganEtAl2019}] \label{def:cbf}
Suppose the dynamics is \eqref{eq:dynamics} and the safe set $\mathcal{C}$ is given by \eqref{eq:safe_set}.
Then, the function $h$ is a control barrier function (CBF) if there exists an extended class $\mathcal{K}_{\infty}$ function $\alpha$ \cite{AmesCooganEtAl2019} such that
\begin{align} \label{eq:CBF}
    \sup_{\mathbf{u} \in \mathcal{U}}\left[L_{f} h(\mathbf{x})+L_{g} h(\mathbf{x})\mathbf{u}+\alpha(h(\mathbf{x}))\right] \geq 0,\ \forall \mathbf{x} \in \mathcal{D},
\end{align}
where $L_X p(x)$ is the Lie derivative of a function $p:\mathcal{P} \rightarrow \mathbb{R}$ with respect of a vector field $X$ at point $x\in \mathcal{P}$.
\end{definition}
We define the set consisting of control inputs that render $\mathcal{C}$ forward invariant as follows,

\vspace{-3mm}
{\small
\begin{align}
    \hspace{-3mm}S_{\mathrm{cbf}}(\mathbf{x}) \coloneqq \left\{\mathbf{u} \in \mathcal{U} \mid L_{f} h(\mathbf{x})+L_{g} h(\mathbf{x})\mathbf{u}+\alpha(h(\mathbf{x})) \geq 0\right\}. 
\end{align}
}
\vspace{-6mm}

Given such a definition, then we have the following theorem regarding the forward invariance of CBFs.
\begin{theorem} [\cite{AmesCooganEtAl2019}] \label{th:CBF_forward_invariance}
Let $\mathcal{C} \subseteq \mathcal{D} \subseteq \mathbb{R}^n $ be the set defined in \eqref{eq:safe_set}. 
If $h$ is a CBF on $D$ and $\frac{\partial h}{\partial \mathbf{x}}(\mathbf{x})\ne 0$ for all $\mathbf{x} \in \partial C$, then any Lipschitz continuous controller $\mathbf{u}:\mathcal{D}\rightarrow\mathcal{U}$ such that $\mathbf{u}\in S_{\mathrm{cbf}}(\mathbf{x})$ will render the set $\mathcal{C}$ forward invariant, that is, the system is guaranteed to be safe.
Moreover, the set $\mathcal{C}$ is asymptotically stable in $\mathcal{D}$.
\end{theorem}

\subsection{Control Lyapunov Function (CLF)} \label{subsec:CLF}

Like CBFs, CLFs offer similar conditions but for stability.
\begin{definition}[Control Lyapunov Function] \label{def:clf}
Given dynamics \eqref{eq:dynamics}, a continuously differentiable function $V:\mathcal{D}\to\mathbb{R}_{\geq0}$ is a control Lyapunov function if it is positive definite and there exists a class $\mathcal{K}$ function $\gamma$ such that
\vspace{-3mm}
\begin{align} \label{eq:CLF}
    \inf_{\mathbf{u} \in \mathcal{U}} \left[L_{f} V(\mathbf{x})+L_{g} V(\mathbf{x})\mathbf{u} + \gamma(V(\mathbf{x}))\right] \leq 0.
\end{align}
\end{definition}
The condition \eqref{eq:CLF} ensures that there exists a control input $\mathbf{u}$ that can decrease the value of $V$ along the system's trajectories.
We define the set consisting of control inputs that stabilize the system as follows,

\vspace{-3mm}
{\small
\begin{align}
    \hspace{-3mm}S_{\mathrm{clf}}(\mathbf{x}) \coloneqq \left\{\mathbf{u} \in \mathcal{U} \mid L_{f} V(\mathbf{x})+L_{g} V(\mathbf{x})\mathbf{u}+\gamma(V(\mathbf{x})) \leq 0\right\}. 
\end{align}
}

Given such a definition, then we have the following theorem regarding stability.

\begin{theorem} \label{th:CLF_stabilization}
Let $V$ be a CLF for the system \eqref{eq:dynamics} on $\mathcal{D}$. 
Then, there exists a Lipschitz continuous controller $\mathbf{u}\in S_{\mathrm{clf}}(\mathbf{x})$  asymptotically stabilizes the system to the origin.\footnote{Equivalently, an equilibrium state.}
\end{theorem}

\section{Conditional Motion Planning with Diffusion}\label{sec:proposed_model}

In this section, we introduce \algo, a conditional diffusion model designed for robot motion planning in dynamic environments.
Our proposed robot planner generates a controller that enables the robot to reach a goal while avoiding collisions with dynamic obstacles (e.g., pedestrians).

\subsection{Problem Setting}

\begin{figure*}[t]
\vspace{2mm}
  \centering
  \includegraphics[width=0.85\linewidth]{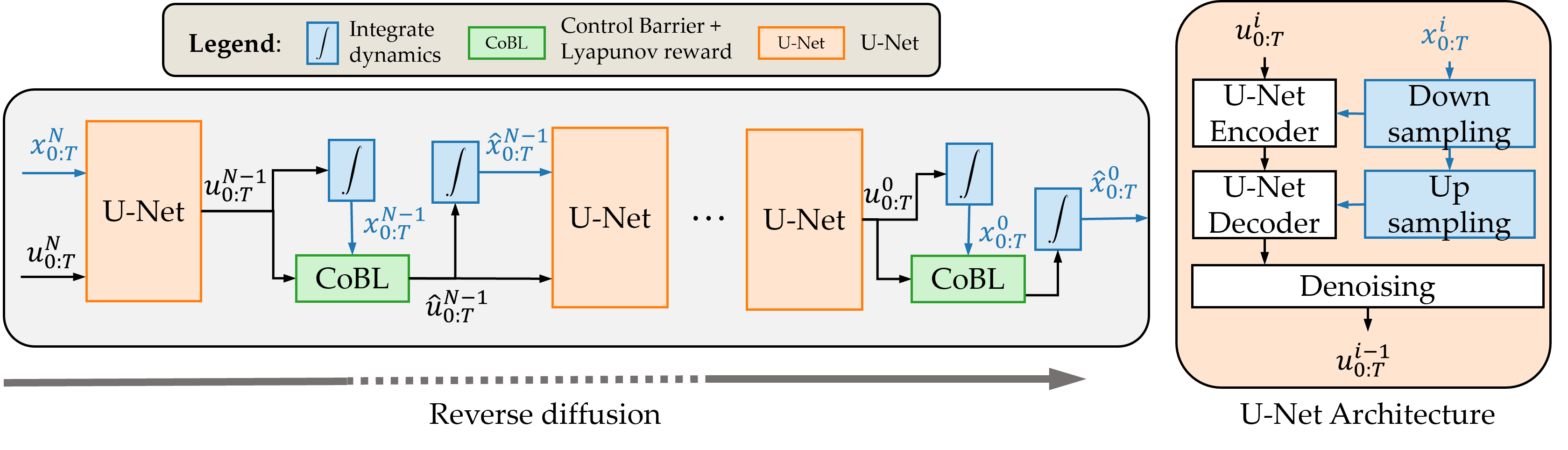}
  \vspace{-2mm}
  \caption{Illustration of planning with CoBL-Diffusion. The left figure depicts the reverse diffusion process of the proposed model. The right figure illustrates the U-Net architecture employed in the proposed model.}
  \label{fig:model_architecture}
  \vspace{-3mm}
\end{figure*}

We denote robot's states as $\mathbf{x}=[\mathbf{x}_0,\ \ldots,\ \mathbf{x}_T] \in \mathcal{D}\subseteq\mathbb{R}^n$ and control inputs as $\mathbf{u}=[\mathbf{u}_1,\ \ldots,\ \mathbf{u}_T] \in \mathcal{U} \subseteq \mathbb{R}^m$, where $T$ represents the time horizon.
Assume that there are $Q$ moving obstacles in the environment and denote the state and control of each obstacle at time $t$ as $\mathbf{x}_{q,t}, \mathbf{u}_{q,t}$ for $q=1,\ldots,Q$.
Consider discrete-time control affine dynamics of the robot:
\begin{align}\label{eq:discrete_system}
    \mathbf{x}_{t+1}=f_d(\mathbf{x}_t, \mathbf{u}_t)
\end{align}
where $f_d:\mathcal{D}\rightarrow\mathcal{D}$ is a continuous function.
For socially acceptable robot navigation, it is essential not only to avoid obstacles but also to generate human-like movements.
Safe and human-like planning in dynamic environments can be formulated as the following trajectory optimization problem:
\begin{align}
    \min_{\mathbf{u}_{0:T}} & \quad J_T(\mathbf{x}_T) + \sum_{t=0}^T J(\mathbf{x}_t, \mathbf{u}_t) \label{eq:prob_objective}\\ 
     \mathrm{s.t.}\ &\quad\mathbf{x}_{t+1} = f_d(\mathbf{x}_t, \mathbf{u}_t) \label{eq:prob_dynamics}\\
    &\quad c(\mathbf{x}_t, \mathbf{x}_{q,t}, \mathbf{u}_t, \mathbf{u}_{q,t}) \geq 0, \: q=1,\ldots,Q \label{eq:prob_safety}\\
    &\quad \mathbf{x}_t \in \mathcal{D},\; \mathbf{u}_t \in \mathcal{U},\; t=0,\ldots,T \label{eq:prob_state_control_constraints}
\end{align}
where $J(\cdot)$ represents a cost function encompassing various planning objectives (e.g., control efficiency, human-likeness, and smoothness), $J_T(\cdot)$ denotes a terminal cost, and $c(\cdot)$ denotes a safety constraint such as avoiding collisions with obstacles.

However, it is nontrivial to specify a general cost function that encodes fluent, human-like, and socially acceptable behaviors; indeed, synthesizing such behavior is a research area itself \cite{SchwartingPiersonEtAl2019,MavrogiannisAlves-OliveraEtAl2021,GeldenbottLeung2024,KretzschmarSpiesEtAl2016}.
Instead, we seek to generate human-like robot control sequences $\mathbf{u}_{0:T}$ via a generative model trained on pedestrian data, while leveraging control-theoretic tools to encode safety constraints and goal objectives, as described by \eqref{eq:prob_objective}--\eqref{eq:prob_state_control_constraints}, to be imposed on the robot.

\subsection{Model Architecture}

Our proposed architecture is inspired by Diffuser \cite{JannerDuEtAl2022}, but there are several key differences. 
The trajectory generation process of \algo ~is shown in Fig.~\ref{fig:model_architecture}

\subsubsection{U-Net Conditioning on Trajectories}
To maintain dynamic consistency, our model generates only control inputs, that is, compared to \eqref{eq:diffuser_traj}, we have $\boldsymbol{\tau}=\mathbf{u}=[\mathbf{u}_1,\ \ldots,\ \mathbf{u}_T]$, and the states are obtained by feeding the controls into the system dynamics \eqref{eq:discrete_system} (see the blue box in Fig.~\ref{fig:model_architecture} left), rather than simultaneously predicting control inputs and states.
While Diffuser conditions the goal by replacing the terminal state, the same inpainting method cannot be applied to \algo ~since it predicts controls only.
Therefore, the states to be satisfied are fed into each resolution of the U-Net \cite{RonnebergerBrox2015} (see orange boxes in Fig.~\ref{fig:model_architecture}) to condition the generated controls for the next denoising step.
During the training of the proposed model, the states obtained by integrating the control inputs corrupted by noise are fed into the U-Net.
This is because the states provided as conditions during the generation process are noisy, obtained by integrating noisy controls in the previous diffusion step with the system dynamics \eqref{eq:discrete_system}.

\subsubsection{Loss Function for Trajectory Error}
Conditioning the U-Net on states is still not sufficient for the diffusion model to generate a controller that reaches the goal.
Therefore, in addition to the conventional diffusion loss $L_\mathrm{simple}$ \eqref{eq:diff_loss} computed on the controls alone, we introduce a loss function $L_{\mathrm{traj}}$ that measures the error between the ground truth and denoised trajectories:
\begin{align}
    L_{\mathrm{traj}} = \frac{1}{T+1}\sum_{t=0}^T\|\mathbf{x}_t-\hat{\mathbf{x}}_t\|^2,
\end{align}
where $x_t$ and $\hat{x}_t$ represent ground truth and generated states at a timestep $t$ respectively.
Here, the generated states are explicitly computed from the denoised control sequence and system dynamics.
This loss function encourages the synthesis of the controller that achieves the given state at each timestep.

\subsubsection{Temporal Weighting for Covariance}
\label{subsubsec:temporal weighting}
The behavior of a generated trajectory depends on the sequence of control inputs by integrating the dynamics.
As a result, errors in control inputs accumulate over time, meaning errors at earlier timesteps can lead to large errors in states later in the horizon.
Therefore, denoising the later control sequence has little effect if the earlier control inputs have not converged. 
To encourage the convergence of the earlier control sequence, the covariances of the forward diffusion \eqref{eq:forward_diff} and reverse diffusion \eqref{eq:reverse_diff} are weighted by $\mathbf{V}$ as follows:
\begin{align} \label{eq:variance_weight}
    \mathbf{V} = \mathrm{diag}(v_0^2, \ldots, v_T^2), 
\end{align}
where $v_0, \ldots, v_T$ are scheduled to increase monotonically with respect to the time steps. 

\subsection{Reward Function} \label{subsec:reward_func}
As introduced in Section~\ref{subsec:diffuser}, a user-defined reward function can be used to guide the reverse diffusion process \eqref{eq:diffuser_reverse_diffusion} to generate outputs with some desirable behavior.
The reverse diffusion process can be guided by multiple reward functions simultaneously, therefore the optimality of the trajectory is denoted using the set of $K$ reward functions $W_k(\mathbf{x}_t, \mathbf{u}_t)$ as $p(O_{t}=1)=\exp{\sum_{k=1}^K W_k(\mathbf{x}_t, \mathbf{u}_t)}$.
Then, we modify the reverse diffusion process as follows:
\begin{align}\label{eq:conditional_reverse_diffusion}
    p_\theta(\mathbf{u}^{i-1}|\mathbf{u}^i, O_{1:T}) \approx \mathcal{N} (\mathbf{u}^{i-1};\boldsymbol{\mu}_\theta(\mathbf{u}^i, i)+\mathbf{g},\ \mathbf{V}\mathbf{\Sigma}^i),
\end{align}
where
\begin{align}
    \mathbf{g} &= \nabla_{\mathbf{u}} \log\ p(O_{1:T}|\mathbf{u})|_{\boldsymbol{u}=\boldsymbol{\mu}_\theta(\mathbf{u}^i, i)}\\
    &= \sum_{k=1}^K\sum_{t=0}^T \nabla_{\mathbf{u}_t} W_k(\mathbf{x}_t, \mathbf{u}_t)|_{\mathbf{u}_t=\boldsymbol{\mu}_\theta(\mathbf{u}^i_t, i)}.
    \vspace{-5mm}
\end{align}
Following \cite{CarvalhoLeEtAl2023}, we drop the scaling factor based on the covariance for the $\mathbf{g}$ shown in \eqref{eq:conditional_reverse_diffusion}.
It addresses the issue of the covariance becoming small towards the end of the reverse diffusion, which would render the guidance ineffective.
Next, we present the CBF and CLF reward functions used for guiding the diffusion process to generate a safe and goal-reaching trajectory.

\subsubsection{Control Barrier Function Reward}
The reward function based on control barrier functions (CBFs) (see Section~\ref{subsec:CBF}) is designed to guide the reverse diffusion process to generate a safe control sequence that avoids collision with dynamic obstacles.
We aim to encourage forward invariance of the safe set in the time domain at each diffusion step.
Although the actual dynamics of the robot is discrete given by \eqref{eq:discrete_system}, we assume the robot and obstacles follow the continuous-time dynamics \eqref{eq:dynamics}.
We define pairwise joint dynamics for the robot and each obstacle $q$:
\begin{align}\label{eq:continuous_joint_system}
    \Dot{\mathbf{X}}_q = F(\mathbf{X}_q)+G(\mathbf{X}_q)\mathbf{U}_q, 
\end{align}
where $\mathbf{X}_q=[\mathbf{x},\ \mathbf{x}_q]^T$ and $\mathbf{U}=[\mathbf{u},\ \mathbf{u}_q]^T$ denote the joint state and control, and $F=[f(\mathbf{x}), f_q(\mathbf{x}_q)]^T,\ G=\mathrm{diag}(g(\mathbf{x}), g_q(\mathbf{x}_q))$.
Suppose the function $h_{\mathrm{cbf}}(\mathbf{X}_q)$ is a CBF to avoid a collision with the obstacle.
From Theorem~\ref{th:CBF_forward_invariance}, control input $\mathbf{u}$ that makes the following reward function $W_{\mathrm{cbf}}$ positive ensures forward invariance of the safe set defined by $h_{\mathrm{cbf}}(\mathbf{X}_q)\geq0$:

\vspace{-3mm}
{\small
\begin{align*}
    W_{\mathrm{cbf}}(\mathbf{X}_q, \mathbf{U}_q) = 
    L_F h_{\mathrm{cbf}}(\mathbf{X}_q) + L_G h_{\mathrm{cbf}}(\mathbf{X}_q)\mathbf{U}_q + \alpha(h_{\mathrm{cbf}}(\mathbf{X}_q)).
\end{align*}
}
Therefore, if the reward function $W_{\mathrm{cbf}}$ is positive for each control action in the entire sequence, collision avoidance is guaranteed.
The gradient of this reward function with respect to the control input $\mathbf{u}$ is computed as follows:
\begin{align*}
    \nabla_{\mathbf{u}} W_{\mathrm{cbf}}(\mathbf{X}_q, \mathbf{U}_q) &= L_G h_{\mathrm{cbf}}(\mathbf{X}_q)[\mathbf{1},\mathbf{0}]^T =  L_g h_{\mathrm{cbf}}(\mathbf{X}_q).
\end{align*}
This gradient does not depend on the control input or dynamics of the obstacle and can be computed as long as the obstacle's state is known.

The trajectories learned and generated by the diffusion model are discrete \eqref{eq:discrete_system}. 
Therefore, it may be more appropriate to use a reward function based on discrete-time CBF \cite{AgrawalSreenath2017, XiongZhaiEtAl2022}.
Therefore, we define the reward function based on discrete-time CBF as follows:
\begin{align*}
    W_{\mathrm{dcbf}}(\mathbf{X}_{q,t}, \mathbf{U}_{q,t}) 
    = \Delta h_{\mathrm{cbf}}(\mathbf{X}_{q,t}, \mathbf{U}_{q,t}) + \alpha(h_{\mathrm{cbf}}(\mathbf{X}_{q,t}))\\
    \mathrm{where}\ \ \Delta h_{\mathrm{cbf}}(\mathbf{X}_{q,t}, \mathbf{U}_{q,t}) = h_{\mathrm{cbf}}(\mathbf{X}_{q,t+1})-h_{\mathrm{cbf}}(\mathbf{X}_{q,t}),
\end{align*}
where $\alpha$ is a class $\mathcal{K}$ function satisfying $\alpha(r)<r$ for all $r>0$.
The performance of continuous-time and discrete-time CBF rewards are compared in Section~\ref{sec:experiment}.

\subsubsection{Control Lyapunov Function Reward}

The reward function based on Control Lyapunov Functions (CLFs) in Section~\ref{subsec:CLF} guides the reverse diffusion process to generate a control sequence that encourages convergence to a given goal $\mathbf{x}_g$.
Similar to CBF reward, we assume that the dynamics of the robot is \eqref{eq:dynamics}.
Suppose $h_{\mathrm{clf}}$ is a CLF, then, according to Theorem~\ref{th:CLF_stabilization}, control inputs $\mathbf{u}$ that make the following reward function $W_{\mathrm{clf}}$ positive will drive the state of the robot $\mathbf{x}$ to $\mathbf{x}_g$:

\vspace{-3mm}
{\small
\begin{align}
    W_{\mathrm{clf}}(\mathbf{x}, \mathbf{u}) =  -L_f h_{\mathrm{clf}}(\mathbf{x}) - L_g h_{\mathrm{clf}}(\mathbf{x})\mathbf{u} - \alpha(h_{\mathrm{clf}}(\mathbf{x}))
\end{align}
}

The gradient of this reward function with respect to the control input $\mathbf{u}$ is computed as follows:
\begin{align}
    \nabla_{\mathbf{u}} W_{\mathrm{clf}}(\mathbf{x}, \mathbf{u}) = -L_g h_{\mathrm{clf}}(\mathbf{x})
\end{align}

Same as the CBF reward, to compare with the continuous-time CLF reward, we define the discrete-time CLF reward:
\begin{align*}
    W_{\mathrm{dclf}}(\mathbf{x}_t, \mathbf{u}_t) = -\Delta h_{\mathrm{clf}}(\mathbf{x}_{t}, \mathbf{u}_t) - \gamma \|\mathbf{x}_{t}-\mathbf{x}_g\|^2\\
    \mathrm{where}\ \ \Delta h_{\mathrm{clf}}(\mathbf{x}_{t}, \mathbf{u}_t) = h_{\mathrm{clf}}(\mathbf{x}_{t+1})-h_{\mathrm{clf}}(\mathbf{x}_t), \: \gamma > 0.
\end{align*}

\subsection{Conditional Motion Planning with Reverse Diffusion}

\begin{figure}[!t]
\vspace{-2mm}
    \begin{algorithm}[H]
        \caption{Conditional Planning with CoBL-Diffusion}
        \label{alg}
        \begin{algorithmic}[1]
        \State Observe start $\mathbf{x}_s$ and goal $\mathbf{x}_g$
        \State initialize controller $\mathbf{u}^N\sim \mathcal{N}(0,\ \mathbf{V}\mathbf{I})$
        \State initialize trajectory $\mathbf{x}^N$
        \State $\mathbf{u}^{N-1}\sim \mathcal{N}(\boldsymbol{\mu}_\theta(\mathbf{u}^N,\ \mathbf{x}^N), \mathbf{V}\mathbf{\Sigma}^N)$
        \For {$i=N-1,\ldots,1$}
        \State Compute $\mathbf{x}^i$ by $\mathbf{x}_s,\ \mathbf{u}^i$ and dynamics 
        \State \small{\texttt{// evaluate the reward functions}}
        \State $\mathbf{g}=\sum_{k=1}^K\sum_{t=0}^T \nabla_{\mathbf{u}^i_t} W_k(\mathbf{x}^i_t, \mathbf{u}^i_t)$
        \State Check $W_k(\mathbf{x}^i_t, \mathbf{u}^i_t)$ for positive and mask $\mathbf{g}$
        \State \small{\texttt{// apply the guidance of rewards}}
        \State $\hat{\mathbf{u}}^i = \mathbf{u}^i + \mathbf{g}$
        \State Compute $\hat{\mathbf{x}}^i$ by $\mathbf{x}_s,\ \hat{\mathbf{u}}^i$ and dynamics 
        \State \small{\texttt{// constrain the start and goal points}}
        \State $\hat{\mathbf{x}}^i_{0},\ \hat{\mathbf{x}}^i_{T} \leftarrow \mathbf{x}_{s},\ \mathbf{x}_{g}$
        \State \small{\texttt{// execute the denoising step}}
        \State $\mathbf{u}^{i-1}\sim \mathcal{N}(\boldsymbol{\mu}_\theta(\hat{\mathbf{u}}^i, \hat{\mathbf{x}}^i),\ \mathbf{V}\mathbf{\Sigma}^i)$
        \EndFor
        \end{algorithmic}
    \end{algorithm}
    \vspace{-10mm}
\end{figure}

In this section, we describe how to guide the reverse diffusion process to generate a safe and goal-reaching controller in dynamic environments by evaluating the trajectory with the reward functions designed in Section~\ref{subsec:reward_func}.
The proposed algorithm is shown in Algorithm~\ref{alg}.

First, we observe start $\mathbf{x}_s$ and goal $\mathbf{x}_g$ of the trajectory, then initialize the trajectory $\mathbf{x}^N$, where $N$ is the number of diffusion steps. 
If there is prior information about the environment, we initialize based on that; otherwise, we simply initialize the trajectory as a straight line connecting the start and goal points.
Next, we sample $\mathbf{u}^{N-1}$ conditioned on $\mathbf{x}^N$ according to the following distribution:
\begin{align}
    \mathbf{u}^{N-1}\sim \mathcal{N}(\boldsymbol{\mu}_\theta(\mathbf{u}^N,\ \mathbf{x}^N),\ \mathbf{V}\mathbf{\Sigma}^N).
\end{align}
Then, the control inputs are fed into the system dynamics \eqref{eq:discrete_system} to obtain the trajectory $\mathbf{x}^{N-1}$ as in line $6$ of Algorithm~\ref{alg}.
At this point, it is important to emphasize that, unlike conventional diffusion model-based trajectory generation, the control sequence and states are consistent.
Then, the trajectory is evaluated by the reward functions defined in Section~\ref{subsec:reward_func}. 
If each reward function is positive at a certain time step, it indicates the trajectory already satisfies each condition at that time.
Therefore, the gradient of the reward function is masked with $0$ at that point.
Before fed the updated states, initial and terminal states are replaced with the given start $\mathbf{x}_s$ and goal $\mathbf{x}_g$ to condition the controls as in line $10$ of Algorithm~\ref{alg}.
Then, the controls $\hat{\mathbf{u}}^i$ are denoised based on the states $\hat{\mathbf{x}}^i$.
This approach aims to iteratively generate controls that gradually satisfy the conditions, instead of employing a common approach of solving a quadratic program (QP) to enforce satisfaction of CBF/CLF constraints which can be computationally expensive over many diffusion and timesteps.
Since there is no need to solve a QP, there is no concern about the feasibility of the trajectories satisfying the constraints.
When using multiple reward functions, the prioritization of conditions can be managed by adjusting their coefficients. 
For example, safety constraints should be emphasized over other constraints, thus necessitating a higher coefficient.

\section{Experiments and Discussion}\label{sec:experiment}

In this section, we validate the effectiveness of the proposed \algo~to synthesize safe and goal-reaching controllers in dynamic environments.
First, we investigate the characteristics of different planners in a single-agent environment. 
Then, we validate the efficacy in a multi-agent environment using the UCY pedestrian dataset \cite{LernerChrysanthouEtAl2007}. 

\subsection{Model Training and Environment Setup}
We trained \algo~using the ETH pedestrian dataset \cite{PellegriniEssEtAl2009}; we trained for 500 epochs on 276,874 8-second trajectories. 
The architecture and hyperparameters of the diffusion model are based on an open-source implementation\footnote{\url{https://github.com/jannerm/diffuser}}.
The Adam optimizer \cite{KingmaBa2015} was used for training with a learning rate set to 2$\times 10^{-5}$ and $\beta = (0.9, 0.999)$.
The covariance weight \eqref{eq:variance_weight} was set to increase linearly from $0.5$ to $1$.
We evaluated the proposed model in two simulation environments created using Trajdata \cite{IvanovicSongEtAl2023}.
The first simulation is a toy environment with a single moving obstacle. 
The second environment is from the UCY pedestrian dataset \cite{LernerChrysanthouEtAl2007} for challenging multi-agent yet feasible settings.

We assumed the robot follows single integrator dynamics, $\Dot{\mathbf{x}} = \mathbf{u}$.
We use the following CBF and CLF to guide the reverse diffusion process:
\begin{align*}
    h_{\mathrm{cbf}}(\mathbf{X}_q) &= (x-x_q)^2 + (y-y_q)^2 - r^2, \\
    h_{\mathrm{clf}}(\mathbf{x}) &= \|\mathbf{x}-\mathbf{x}_g\|^2,
\end{align*}
where $\mathbf{x}=[x, y]$, $\mathbf{x}_q=[x_q, y_q]$ represent the states of the robot and the $q$-th obstacle respectively, $r$ is a barrier radius, and $\mathbf{x}_g=[x_g, y_g]$ is a given goal point.

\subsection{Comparison Methods and Evaluation Metrics}
\label{subsec:comparison_methods}

We compare \textbf{CoBL-Diffusion (CoBL)} with the baseline method \textbf{CBF-QP}, which solves a QP with CBF at every timestep given a nominal straight line trajectory to the goal, \textbf{Velocity Obstacle (VO)} \cite{FioriniShiller1998}, which is a geometric approach for collision avoidance that uses the relative velocity between a robot and an obstacle to identify potential collision velocities, and \textbf{Diffuser} \cite{JannerDuEtAl2022}, which is a planning with a diffusion model and the generated plan is modified by CBF-QP to ensure collision avoidance.
Variants of the proposed CoBL-Diffusion are also investigated: \textbf{dCoBL-Diffusion (dCoBL)}, which uses discrete-time CBF and CLF rewards; \textbf{CoB-Diffusion (CoB)}, which only uses a CBF reward; 
\textbf{DCoL-Diffusion (DCoL)}, which uses CLF and a distance reward $W_{\mathrm{dist}}$ instead of a CBF reward:
\begin{align*}
    W_{\mathrm{dist}}(\mathbf{x}_t, \mathbf{u}_t) = \|\mathbf{x}_t-\mathbf{x}_{q,t}\|^2-r^2,
\end{align*}
where $r$ is a barrier radius.
Comparing the CBF reward, the distance reward considers only the distance to obstacles and ignores the dynamics.
The coefficients for (discrete-time) CBF and CLF rewards are fixed to $0.3$ and $0.1$, and for distance reward is fixed to $0.3$.
In this experiment, we set the barrier radius to $1\ m$ and the collision radius to $1\ m$.
For a clear comparison, the reward functions for collision avoidance consider only the nearest obstacle, although considering multiple obstacles could potentially improve performance and is deferred for future work.

Each method is evaluated based on three metrics: collision rate, goal-reaching, and smoothness.
The collision rate is calculated as the ratio of simulations that violate the collision radius among all simulations.
The goal-reaching is evaluated by the root squared error between the goal and the final point. 
The smoothness is measured by the worst-case root squared difference in control inputs (in these experiments, velocity).

\subsection{Safe Planning in Single-Agent Environment} \label{subsec:sim_simple_environment}

\begin{table*}[t]
  \vspace{5mm}
  \caption{Simulation results for the single and multi-agent environments.}
  \label{tab:quantitative comparison}
  \vspace{-4mm}
  \begin{center}
    {\small
     \begin{tabular}{|c|cc|ccc|ccc|}
     \hline
     Planner & \multicolumn{2}{c|}{Reward} & \multicolumn{3}{c|}{\textbf{Single-Agent environment}}  & \multicolumn{3}{c|}{\textbf{Multi-Agent environment}}\\
     \hline
     & Safety & Goal & Coll. ($\downarrow$) & Goal-Reach. ($\downarrow$) & Smoothness ($\downarrow$) &  Coll. ($\downarrow$) & Goal-Reach. ($\downarrow$) & Smoothness ($\downarrow$)\\ \hline
     \rowcolor{blond}
     \textbf{CoBL}& $W_\mathrm{CBF}$ & $W_\mathrm{CLF}$ & $\mathbf{0}$ \% & $\mathbf{0.20 \pm 0.10}$ & $\mathbf{0.45 \pm 0.16}$ & $\mathbf{0.5}$ \% & $\mathbf{0.41 \pm 0.20}$ & $\mathbf{0.65 \pm 0.31}$ \\ 
     CoB  & $W_\mathrm{CBF}$ & -- & $6$ \% & $1.36 \pm 0.51$ & $2.01 \pm 0.19$  & $20.0$ \% & $3.60 \pm 1.74$ & $1.66 \pm 0.45$  \\
     dCoBL  & $W_\mathrm{dCBF}$ & $W_\mathrm{dCLF}$ & $8$ \% & $0.18 \pm 0.08$ & $0.53 \pm 0.15$  & $48.0$ \% & $0.38 \pm 0.16$ & $0.45 \pm 0.14$  \\
     DCoL  & $W_\mathrm{dist}$ & $W_\mathrm{CLF}$  &  $4$ \% & $0.13 \pm 0.06$ & $0.60 \pm 0.24$  & $42.5$ \% & $1.10 \pm 1.89$ & $1.52 \pm 1.90$  \\
    CoBL$^-$ & $W_\mathrm{CBF}$ & $W_\mathrm{CLF}$ & -- & -- & -- & $8.5$ \% & $0.48 \pm 0.34$ & $0.73 \pm 0.39$  \\
     \hline
     CBF-QP & -- & -- & $0$ \% & $0.13$ & $10.78$  & $62.5$ \% & $0.07 \pm 0.02$ & $4.98 \pm 4.20$  \\
     VO & -- & -- & $0$ \% & $0.33$ & $2.00$  & $0$ \% & $0.31 \pm 0.54$ & $1.78 \pm 0.32$  \\
     Diffuser & -- & -- & $0$ \% & $2.06 \pm 0.49$ & $0.78 \pm 0.04$  & $29$ \% & $7.94 \pm 3.22$ & $1.74 \pm 0.31$  \\
     \hline
     Human & -- & -- & -- & -- & -- & $74.5$ \%  & $0$ & $0.08 \pm 0.04$\\ \hline
     \end{tabular}
    }
    \end{center}
    \vspace{-4mm}
\end{table*}

\begin{figure}[t]
\centering
\begin{minipage}[t]{0.3\linewidth}
  \centering
  \includegraphics[width=\linewidth]{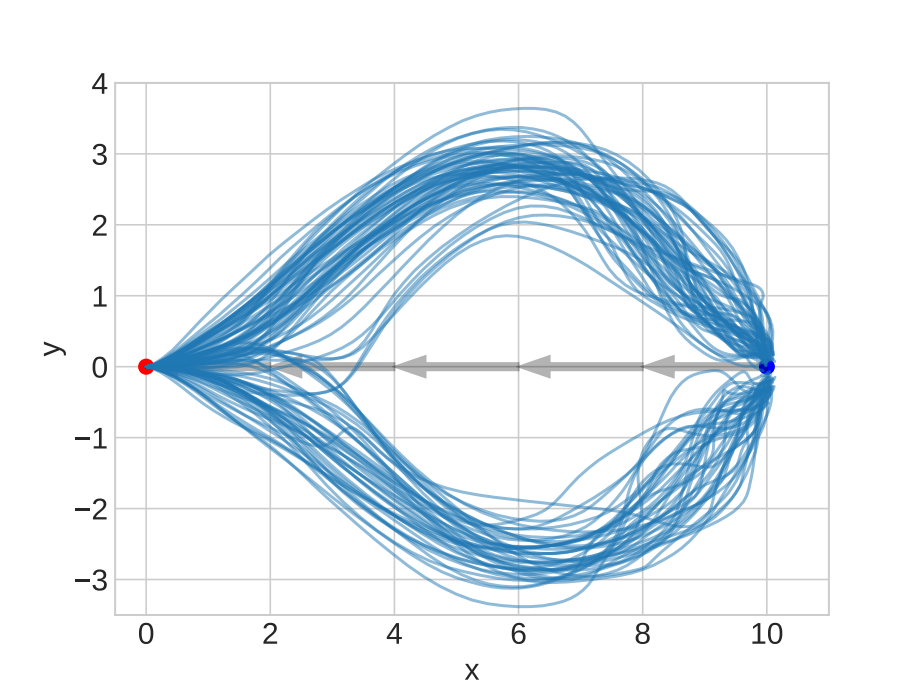}
  \subcaption{CoBL-Diffusion.}
\end{minipage}
\begin{minipage}[t]{0.3\linewidth}
  \centering
  \includegraphics[width=\linewidth]{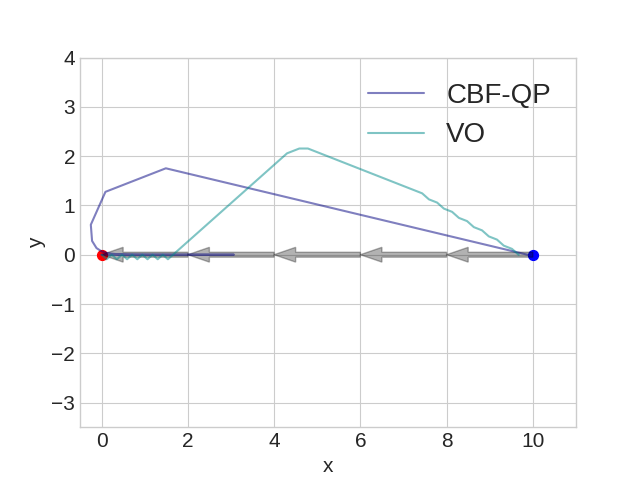}
  \subcaption{CBF-QP and VO.}
\end{minipage}
\begin{minipage}[t]{0.3\linewidth}
  \centering
  \includegraphics[width=\linewidth]{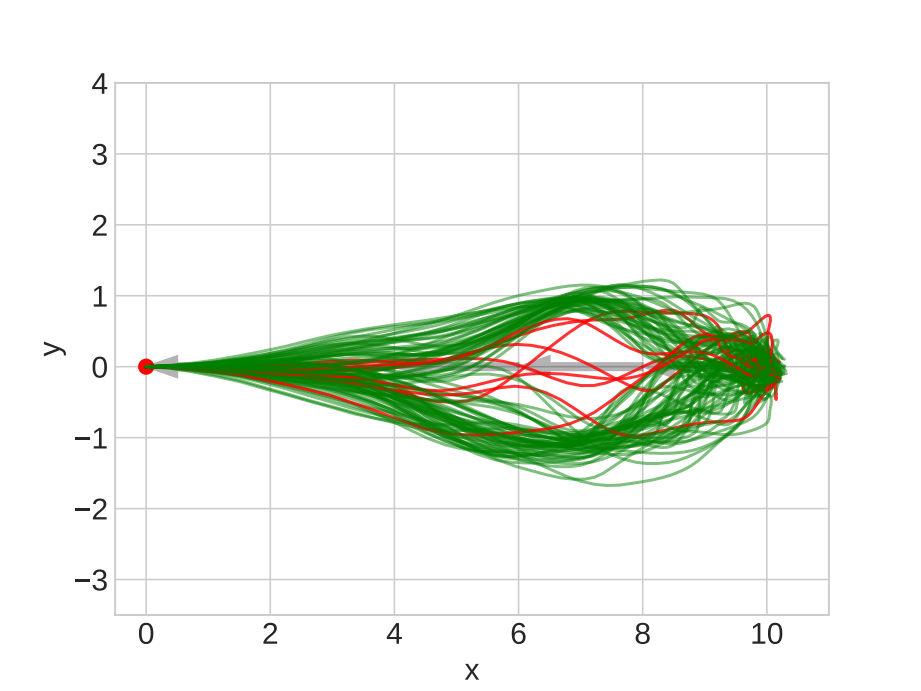}
  \subcaption{\centering dCoBL-Diffusion.}
\end{minipage}\\
\vspace{4mm}
\begin{minipage}[t]{0.3\linewidth}
  \centering
  \includegraphics[width=\linewidth]{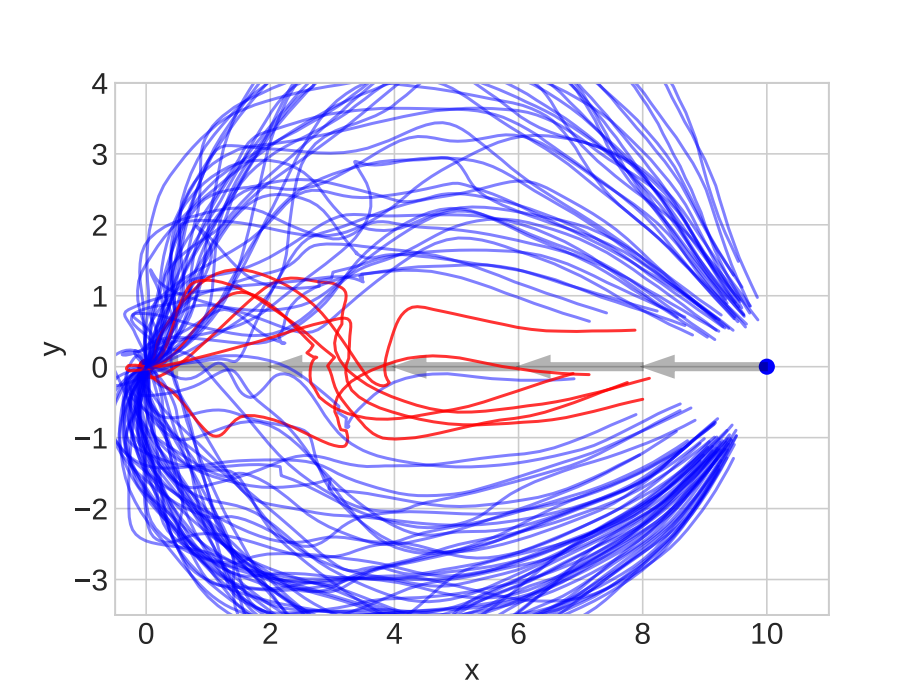}
  \subcaption{CoB-Diffusion.}
\end{minipage}
\begin{minipage}[t]{0.3\linewidth}
  \centering
  \includegraphics[width=\linewidth]{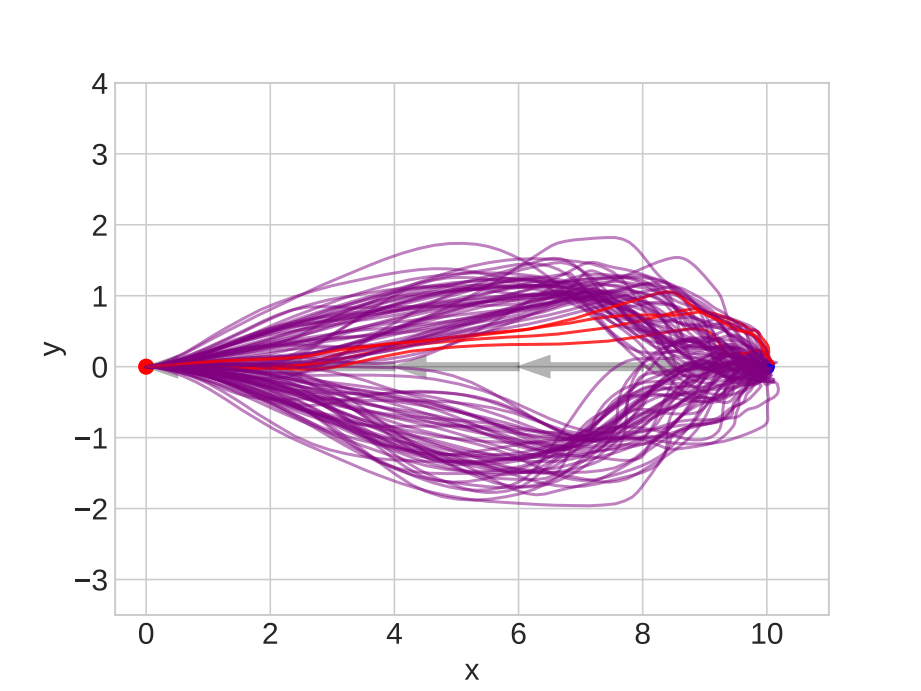}
  \subcaption{DCoL-Diffusion.}
\end{minipage}
\begin{minipage}[t]{0.3\linewidth}
  \centering
  \includegraphics[width=\linewidth]{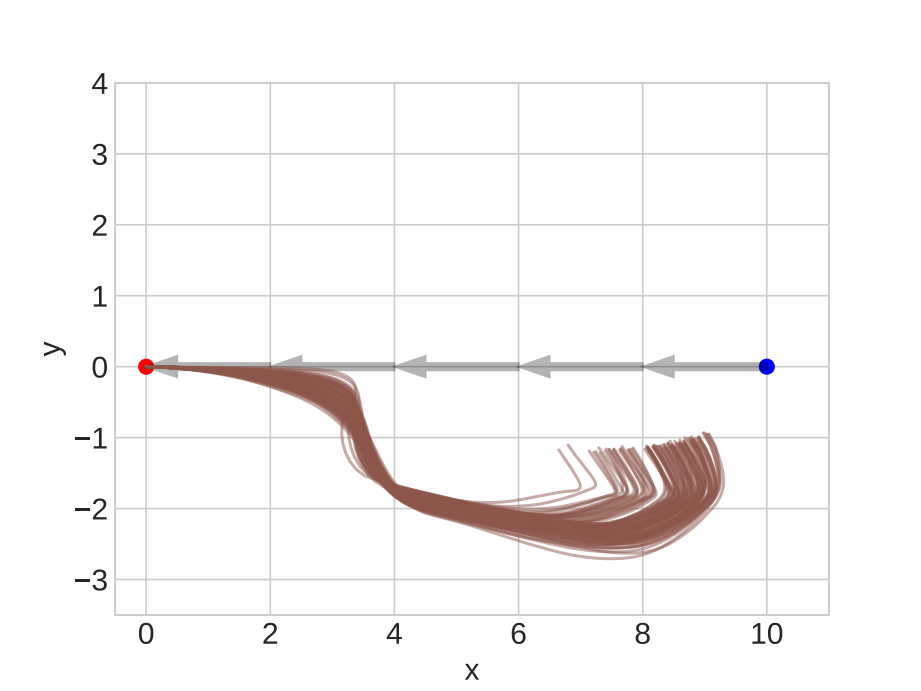}
  \subcaption{Diffuser.}
\end{minipage}
\vspace{2mm}
\caption{Generated trajectories in the single-agent environment with various planning methods. Red trajectories experience collisions.}
\label{fig:ex_toy_traj}
\vspace{-3mm}
\end{figure}

We evaluate the diffusion model guided by each reward function in a single-agent environment.
The environment contains an obstacle moving from location $(10,0)$ to $(0,0)$ (right to left).
The robot aims to reach location $(10,0)$ from $(0,0)$ (left to right) while avoiding the oncoming obstacle.
The position and velocity of the obstacle are provided in advance for generating a plan.
The initial trajectory given at the beginning of reverse diffusion is a straight line connecting the given start and goal points, and is therefore infeasible as it collides with the obstacle. 

The results of simulations are summarized in Table~\ref{tab:quantitative comparison}, with the trajectories illustrated in Fig.~\ref{fig:ex_toy_traj}.
Note that the CBF-QP and VO methods are deterministic.
The CoBL successfully generated collision-free and smooth plans. 
While the CBF-QP and VO methods successfully avoided obstacles, they produced less smooth trajectories, with VO also exhibiting the poor goal-reaching performance.
Although the Diffuser avoided collisions with CBF-QP, it resulted in the worst deviation from the goal.
Ablation studies revealed that using discrete-time rewards to adapt to system dynamics resulted in a decline in the safety of the generated plans.
In terms of goal-reaching performance, CoBL was comparable with the other CoBL variants; DCoL achieved the highest goal-reaching performance but was outperformed by CoBL in terms of collision avoidance and smoothness.
Fig.~\ref{fig:ex_toy_traj} indicates that dCoBL tends to generate more aggressive trajectories, likely due to the stronger influence of the CLF in the discrete-time formulation.
Comparing CoBL and DCoL, using a reward function derived from CBF ensures stronger safety than merely considering the distance from obstacles.
When comparing CoBL with CoB, the contribution of the CLF reward to the goal convergence performance is not evident in the simple environment.
Even without the CLF reward, the proposed diffusion model's ability to synthesize a controller to achieve the given states allows it to reach the goal.
We next investigate \algo~in a dynamic multi-agent scenario.

\subsection{Safe Planning in Real Crowd Environment} \label{subsec:real_ex}

\begin{figure}[t]
  \centering
  \vspace{-5mm}
  \includegraphics[width=0.7\linewidth]{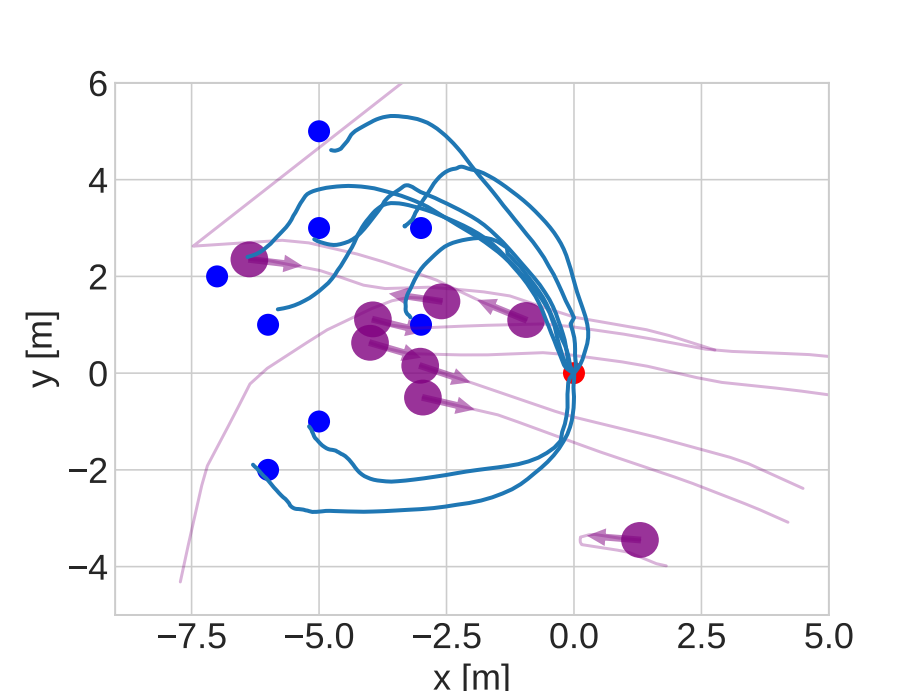}
  \vspace{-1mm}
  \caption{Visualization of the planning environment in a realistic dynamic setting with pedestrians. The initial position of the robot is marked by a red dot, and the target goals are indicated by blue dots. The generated trajectories are represented by blue lines, while the purple circles and lines represent the humans and their respective trajectories.}
  \label{fig:ex_real_environment}
  \vspace{-4mm}
\end{figure}

In this section, we evaluate the performance of our proposed model in a multi-agent environment from a real-world pedestrian dataset. 
We selected a scene from the UCY pedestrian dataset \cite{LernerChrysanthouEtAl2007} that provides a challenging yet comprehensible environment for safe trajectory generation.
The environment used for the simulation has eight pedestrians, as shown in Fig.~\ref{fig:ex_real_environment}.
We selected eight goal locations and ran $200$ simulations, $25$ for each goal.

\noindent\textbf{Full trajectory knowledge.} First, we assume that the robot knows the trajectories of all humans in the field for the entire horizon.
In this setting, the coefficients of CBF, CLF, and Distance reward were set to $0.3$, $0.01$, and $0.3$, respectively to emphasize safety constraints since the number of obstacles increases and the complexity of movement becomes more challenging for ensuring safety.
The CBF barrier radius was set to $1\ m$, providing a buffer for a collision radius of $0.7\ m$.
We included \textbf{CoBL-Diffusion$^-$ (CoBL$^-$)} without covariance weight in our evaluation to investigate the effect of the weight on covariance \eqref{eq:variance_weight} in encouraging convergence of the earlier control sequence.
The results of each model are summarized in Table~\ref{tab:quantitative comparison}.
For reference, the actual 200 pedestrian data with $3-8$ $m$ movements are shown as Human.
The high collision rate of Human is attributed to the defined collision radius being larger than the actual one.

CoBL demonstrated its ability to generate safe plans even in complex environments with multiple moving people, with an average generation time of $4.8$ seconds.
On the other hand, CBF-QP is unable to avoid moving obstacles in a dynamic multi-agent environment.
Although the trajectories generated by VO can safely navigate around multiple obstacles, they often lack smoothness and may become impractical or unpredictable for surrounding humans.
The result of the Diffuser has the worst distance from the goal because they are modified by the CBF-QP after generating a plan.
In contrast, our model integrates collision avoidance into the planning process, enabling the synthesis of safe trajectories from the outset.
Similar to simulations in the single-agent environment, using dCoBL or DCoL rewards makes it challenging to ensure safety.
Comparing CoBL with CoB, while equivalent quality trajectories were generated in the simple environment, the contribution of CLF reward to goal-reaching is more prominent in this more complex setting; the trajectory is constantly modified by the CBF reward, leading to deviations from the goal without the CLF reward.
When comparing the CoBL with the CoBL$^-$, it seems that the temporal weighting introduced for the covariance \eqref{eq:variance_weight} to promote convergence of the earlier control sequence slightly reduces the collision rate; these results prompt further investigation to better understand the impact of the temporal weighting.

\begin{table}[t]
  \caption{Simulation results for varying levels of knowledge in the multi-agent environment.}
  \label{tab:ex_fut}
  \vspace{-4mm}
  \begin{center}
    {\small
     \begin{tabular}{|c|ccc|}
     \hline
     Scenario & Collision & Goal-Reaching & Smoothness \\
     \hline
     FullTraj & $1$ \% & $0.41 \pm 0.20$ & $0.74 \pm 0.34$ \\
     FourSecTraj & $0.5$ \% & $0.40 \pm 0.20$ & $0.68 \pm 0.29$  \\
     TwoSecTraj & $4$ \% & $0.36 \pm 0.20$ & $0.67 \pm 0.32$  \\
     InitTraj & $19.5$ \% & $0.45 \pm 0.30$ & $0.67 \pm 0.32$  \\
     \hline
     \end{tabular}
    }
    \end{center}
    \vspace{-9mm}
\end{table}

\noindent\textbf{Partial trajectory knowledge.}
Next, we consider a more realistic scenario where the robot only knows the past trajectories of the humans in the field. 
In this simulation, we make a constant velocity assumption---obstacles continue to move straight with their last known position and velocity.
The results are shown in Table~\ref{tab:ex_fut} for the following scenarios: FullTraj, where the robot knows the entire 8-second trajectories of the obstacles; FourSecTraj, where only the first 4 seconds of the trajectory is known (and a constant velocity assumption is made after 4 seconds); TwoSecTraj, where only the first 2-seconds of the trajectory is known; and InitTraj, where only the current position and velocity of the obstacles.
We see that for CoBL, it is found that assuming the future motion of obstacles as constant linear motion does not significantly compromise safety guarantees---we still outperform the other baseline methods and CoBL variants from Table~\ref{tab:quantitative comparison}.
It is important to note that in this scene, the obstacles were humans and approximating their movement as constant linear was adequate for a short period.
It is expected that combining our model with more advanced motion prediction models can ensure stronger safety, and we plan on investigating this in future work.

\section{Conclusion and Future Directions}\label{sec:conclusion}

Our proposed \algo~synthesizes robot controllers for safe planning in dynamic multi-agent environments.
Algorithmically, \algo~guides the reverse diffusion process with reward functions based on CBF and CLF.
During the denoising process, \algo~iteratively improves the control sequence towards satisfying the safety and stability constraints.
To evaluate the efficacy of the proposed model for safe planning in dynamic environments, we compared it with existing methods and variants of our proposal.
We found that \algo~can smoothly reach goal locations within an acceptable distance while having very low collision rates.
However, the collision rates increased as the uncertainty in the behaviors of obstacles increased.
These results prompt the need for further investigation in methods to incorporate behavior prediction within the diffusion model. 
Additionally, to promote real-time applications, we plan to explore how to reduce the inference time of our models, such as leveraging ideas from DDIM \cite{SongMengEtAl2021} and consistency models \cite{SongDhariwalEtAl2023}, and investigate the effectiveness of our conditioning approach with them.

\bibliographystyle{IEEEtran}
\bibliography{main,ctrl_papers}

\end{document}